\documentclass{ecai} 
\usepackage{multirow}
\usepackage{soul}

\usepackage{latexsym}
\usepackage{amssymb}
\usepackage{amsmath}
\usepackage{amsthm}
\usepackage{booktabs}
\usepackage{enumitem}
\usepackage{graphicx}
\usepackage{color}
\usepackage{soul}

\newcommand{\BibTeX}{B\kern-.05em{\sc i\kern-.025em b}\kern-.08em\TeX}

\newcommand{\manu}[1]{#1}
\newcommand{\mari}[1]{#1}

\begin{document}


\begin{frontmatter}


\paperid{1036} 


\title{Evaluating the Simulation of Human Personality-Driven Susceptibility to Misinformation with LLMs}







\author[A,B]{\fnms{Manuel}~\snm{Pratelli}\orcid{0000-0002-9978-791X}\thanks{Corresponding Author. Email: manuel.pratelli@iit.cnr.it.}\footnote{Equal contribution.}}
\author[A,B]{\fnms{Marinella}~\snm{Petrocchi}\orcid{0000-0003-0591-877X}\footnotemark}

\address[A]{IIT-CNR}
\address[B]{IMT School for Advanced Studies Lucca}


\begin{abstract}

Large language models (LLMs) make it possible to generate synthetic behavioural data at scale, offering an ethical and low-cost alternative to human experiments. Whether such data can faithfully capture psychological differences driven by personality traits, however, remains an open question.
We evaluate the capacity of LLM agents, conditioned on Big-Five profiles, to reproduce personality-based variation in susceptibility to misinformation, focusing on news discernment, the ability to judge true headlines as true and false headlines as false. Leveraging published datasets in which human participants with known personality profiles rated headline accuracy, we create matching LLM agents and compare their responses to the original human patterns.
Certain trait–misinformation associations, notably those involving Agreeableness and Conscientiousness, are reliably replicated, whereas others diverge, revealing systematic biases in how LLMs internalize and express personality. The results underscore both the promise and the limits of personality-aligned LLMs for behavioral simulation, and offer new insight into modeling cognitive diversity in artificial agents.
\end{abstract}

\end{frontmatter}


\section{Introduction}
News discernment—the ability to judge true news as true and false news as false—has become a core construct in misinformation research. A recent meta-analysis aggregating 300 + effect sizes from more than 60 studies shows that people reliably rate genuine headlines as more accurate than fabricated ones, yet the size of this “accuracy gap” varies widely across individuals \cite{Pfander2025}. Personality traits account for part of this variability, shaping both susceptibility to and sharing of misinformation \cite{calvillo2021personality}.

In a standard experimental paradigm, participants complete a validated inventory to assess their personality according to,  e.g., the Big-Five model \cite{fiske1949consistency,goldberg1990alternative,mccrae1994big,peabody1989some} and then rate the accuracy of a set of headlines \cite{calvillo2021personality,pennycook2019lazy}. Although informative, such large-scale behavioral studies are costly and time-consuming \cite{10.1525/collabra.25293}, and repeated exposure to deceptive content raises ethical concerns about potential harm and data protection \cite{MURPHY2023101713}.

Recent work therefore turns to large language models (LLMs) as a source of rich, low-cost synthetic data \cite{Argyle_Busby_Fulda_Gubler_Rytting_Wingate_2023,long-etal-2024-llms}. Huang et al. \cite{huang2024designing} show that LLM agents can be assigned stable Big-Five profiles, an approach that leverages the shared linguistic basis of both LLMs and lexical personality theory.

\paragraph{Present study.} We investigate whether personality-aligned LLM agents can reproduce the association between  personality trait and news discernment documented in human samples. Positive evidence would offer an ethically lightweight, scalable test-bed for future work on information resilience.

\begin{figure*}[h]
\centering
\includegraphics[width=.45\linewidth]{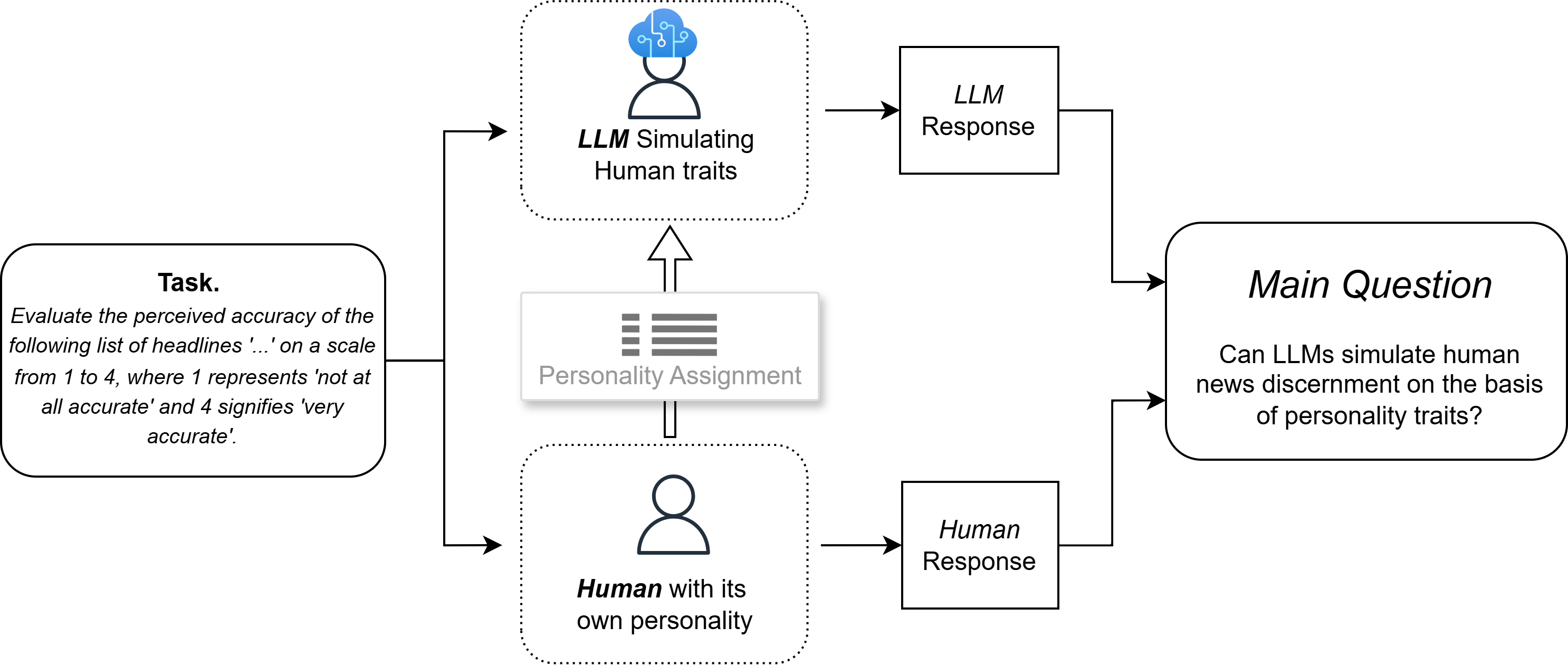}
\caption{Core concept}
\label{fig:idea}
\end{figure*}



To operationalize the study (Figure \ref{fig:idea}), we instantiate one synthetic agent per human participant, feeding each agent the participant’s Big-Five profile via the personality-to-agent pipeline of \citet{huang2024designing}. Every agent receives the identical set of news headlines as in \cite{calvillo2021personality} and rates their accuracy on the same scale used with humans.

Using the same metrics and statistical tools as existing studies, we compare human responses with those of their personality-matched synthetic agents, and investigate whether similar associations between personality traits and news discernment emerge in both groups. Convergent patterns would indicate that personality-aligned LLMs can, at least in part, reproduce the way humans evaluate true versus false news.

\paragraph{Research questions.} The main research question addressed in this work is:

\begin{itemize}
    \item \textbf{RQ\(_{\text{main}}\)} Can large language models, when endowed with explicit Big-Five profiles, simulate human news discernment? 
    \end{itemize}

We decompose it into three empirical sub-questions:

\begin{itemize}
    \item \textbf{RQ1} 
    Does assigning a Big-Five profile to an LLM change its perceived-accuracy ratings for true and false headlines relative to a neutral (no-personality) baseline?
    
    \item \textbf{RQ2} 
    Do personality-conditioned LLMs reproduce the trait–news discernment correlations observed in human samples (i.e., the difference in accuracy ratings between true and false headlines)?

    \item \textbf{RQ3} 
    Do those same agents also mirror human trait effects on belief in misinformation, defined here as the accuracy ratings assigned to false headlines only?
    
\end{itemize}

Building on the personality-to-agent pipeline of \citet{huang2024designing}, we instantiate a suite of LLM agents endowed with explicit Big-Five profiles. To test the robustness of any personality effects, we systematically vary the model parameters (version and temperature), the personality inventory (BFI-2 vs.\ BFI-2-S), the response format (Likert vs.\ Expanded), and
the personality profile source (participant datasets from \citealp{huang2024designing} and \citealp{calvillo2021personality}).

The agents’ accuracy ratings are compared with human data. We use \citet{calvillo2021personality} as our primary benchmark because it employs the same personality profiles and headline set, and we situate our results within the broader evidence synthesised by \citet{calvillo2024personality}, a recent review of studies involving human participants.


\paragraph{Results.} 
Our experiments yield three main findings. 
First, personality conditioning produces statistically significant shifts in LLM accuracy ratings, confirming that injected traits measurably affect news discernment.  
Second, GPT-4o reproduces several human trait–discernment associations: Agreeableness, Conscientiousness, and Open-Mindedness correlate positively with news discernment in both human and synthetic data, whereas patterns for Extraversion and Negative Emotionality diverge, indicating incomplete psychological fidelity.  
Third, when the analysis is restricted to false headlines, personality-aligned LLMs again echo humans—higher Agreeableness and Conscientiousness predict lower \manu{susceptibility} to misinformation—yet the link with Open-Mindedness remains inconsistent across model settings, mirroring mixed evidence in the literature.


\paragraph{Contributions.}
\begin{enumerate}
    \item \textbf{First empirical test of personality–misinformation simulation.}  
    We present, to the best of our knowledge, the first study that uses large language models to replicate the trait–news discernment associations previously observed in human participants.

    \item \textbf{Robustness analysis across inventories and model settings.}  We systematically test two Big-Five inventories, two response scales, and two LLM temperatures, providing a robustness map of personality effects in synthetic agents.

    \item \textbf{Open resources for transparency and reproducibility.}  
    In case of manuscript acceptance, we will release code, prompts, and synthetic-versus-human comparison datasets to enable verification, reuse, and extension by the research community.
\end{enumerate}

\section{Related Work}

\paragraph{Human Personality Traits Assessment.}

Personality traits are stable individual differences influencing cognition, behavior, and social interactions, and are robust predictors of life outcomes such as educational attainment, job performance, and socio-economic status~\cite{soldz1999big,soto2019replicable}. The Big Five model, rooted in the Lexical Hypothesis~\cite{allport1936trait}, emerged through factor-analytic studies~\cite{norman1963toward, norman1967personality, wiggins1979personality}, identifying five broad dimensions: Open Mindedness, Conscientiousness, Extraversion, Agreeableness, and Negative Emotionality~\cite{fiske1949consistency, goldberg1990alternative, mccrae1994big, peabody1989some}. 

Big-Five inventories trade length for precision: the 10-item TIPI~\cite{gosling2003very,ahmed2022personality} enables large surveys, the 41-item IPIP~\cite{buchanan2005implementing} yields richer profiles, and the 60-/30-item BFI-2 and BFI-2-S refine psychometrics and update labels~\cite{soto2017next,soto2017short}. To curb acquiescence bias and boost reliability, Zhang et al.~\cite{zhang2024improving} recently released three revised BFI-2 versions (Expanded, Item-Specific-Full, Item-Specific-Light), widening their use in both human and AI-centered personality research.

\paragraph{Links between Human Traits and Misinformation.}

Given the intrinsic interplay between individual characteristics and online behavior, a growing body of work explores how psychological and demographic factors shape individuals’ \textit{susceptibility to misinformation}, namely, their tendency to believe, share, or fail to detect false or misleading content. Predictors include personality traits (e.g., Big Five, HEXACO, Dark Triad), cognitive abilities, and sociodemographic variables such as age, gender, education, and political ideology~\cite{pennycook2019lazy, calvillo2021personality, ahmed2022social, buchanan2021trust, barman2021exploring, calvillo2024personality, peter2024role,sultan2024susceptibility,barman2024discerning}.

In the present study, we focus on links between personality traits, assessed using the Big Five model, and 
susceptibility to misinformation. We consider two behavioral outcomes: (i) \textit{belief in misinformation}, measured as the perceived accuracy of false headlines, and (ii) \textit{news discernment}, defined as the difference in perceived accuracy between true and false headlines~\cite{pennycook2019lazy}.

Three studies have investigated Big Five correlates of news discernment using both true and false headlines. Calvillo et al.~\cite{calvillo2021personality} used the BFI-2-S with 353 U.S. participants, evaluating 12 true and 12 false political headlines. Discernment was negatively associated with Extraversion and positively with Agreeableness, Conscientiousness, and Openness. Sindermann et al.~\cite{sindermann2021evaluation} replicated the negative association with Extraversion in a German sample (n=530), with no other significant effects. Peter et al.~\cite{peter2024role}, using HEXACO and Dark Triad models, found that lower discernment was linked to lower Conscientiousness, Openness, and Honesty-Humility, and to higher narcissism, Machiavellianism, and psychopathy.

Other studies have assessed susceptibility to misinformation using only false headlines. Ahmed et al.~\cite{ahmed2022social}, using a 10-item BFI in Singapore (n=500), found positive associations with Openness, Extraversion, and Negative Emotionality and a negative one with Conscientiousness. In a U.S. sample (n=750), Ahmed et al.~\cite{ahmed2022personality} showed that Extraversion predicted belief in pro-conservative misinformation, while Agreeableness predicted lower belief. Buchanan et al.~\cite{buchanan2021trust} (UK, n=172) found no significant associations. Shephard et al.~\cite{shephard2023everyday} examined Emotional Stability among undergraduates but found no link to accuracy judgments.

Table~\ref{tab:big5_misinfo_summary} summarizes findings from these studies, offering a comparison for our results.

\paragraph{Personality and LLMs.}


Large Language Models (LLMs) are increasingly employed to simulate human behavior across various research domains~\cite{pratelli2024evaluation,mei2024turing,huang2024designing,xi2025rise,xu2024ai}. This growing use has prompted investigations into their intrinsic biases and the extent to which they naturally exhibit, or can be conditioned to emulate—human, like personality traits.

Some studies aim to characterize the \textit{intrinsic personalities} of LLMs~\cite{la2024open, frisch2024llm}, while others explore the feasibility of \textit{assigning} personality traits to LLMs in order to reproduce human behavioral patterns and decision-making styles~\cite{huang2024designing, giorgi2024human, mei2024turing, kozlowski2024simulating}.

La Cava e Tagarelli~\cite{la2024open} evaluated intrinsic personality traits across multiple open-source LLMs and assessed their responsiveness to personality-conditioned prompts. The authors found that: (i) different LLMs exhibit distinct personality profiles; (ii) conditioning LLMs with specific personality traits has mixed success: most models retain their intrinsic biases and fail to fully emulate the imposed personality, a phenomenon described as “closed-mindedness”; and (iii) combining personality traits with role-based prompting improves alignment with target profiles.

Giorgi et al.~\cite{giorgi2024human} compared the biases of persona-conditioned LLMs with those of human annotators in hate speech detection. While persona-based LLMs did exhibit identifiable biases, these differed significantly from those of human participants, revealing important design considerations for human-AI collaboration in annotation tasks.

Kozlowski et al.~\cite{kozlowski2024simulating} highlighted the potential of LLMs to simulate culturally and socially grounded human subjects. They proposed a methodological foundation for simulating human participants, identifying limitations such as atemporality, uniformity, and social desirability bias. The authors advocate for an ongoing methodological program to keep pace with advances in model capabilities.

Mei et al.~\cite{mei2024turing} evaluated whether LLMs could pass a behavioral Turing Test using classic economic and psychological tasks. ChatGPT-4 exhibited behavioral and personality patterns statistically indistinguishable from human data collected across 50 countries. Interestingly, while LLMs demonstrated adaptability to context and framing, their behaviors were often more altruistic and cooperative than those of typical human participants.

Huang et al.~\cite{huang2024designing} advanced this line of research by assigning psychometrically valid personality profiles to LLM agents using the Big Five model~\cite{soto2017next}. Based on the \textit{lexical hypothesis} (see previous paragraph), they first established that LLMs can represent personality constructs in semantic space, and then validated their findings through simulated and human-aligned responses to personality tests.  In subsequent experiments, personality-assigned agents made decisions in risk-taking and ethical dilemma scenarios. While the results for risk-taking were consistent with known human patterns, the agents were less consistent with human data in ethical contexts.  However, this psychometric conditioning framework has not yet been applied to misinformation scenarios, leaving open the question of whether such personality-engineered agents can mimic human vulnerability or resilience to misinformation.

Salecha et al.~\cite{salecha2024large} found a consistent tendency in large language models (LLMs) to favor socially desirable personality traits (e.g., greater extroversion, less negative emotionality). This raises concerns about the validity of assigning personality traits to LLMs, particularly in cross-cultural or adversarial contexts.

Despite the increasing focus on personality modeling in LLMs, to our knowledge, no previous work has investigated whether personality-matched LLMs exhibit human-like vulnerabilities or resistances to misinformation. The present study aims to fill this gap by investigating whether LLMs, when conditioned to reflect different personality profiles, exhibit discernible patterns of susceptibility to misinformation - mirroring findings from psychological studies of human participants.

\section{Methodology}\label{sec:methodology}

\paragraph{News Discernment and Belief in Misinformation.}

\textit{Susceptibility to misinformation} is typically gauged through two related metrics.  
\textit{Belief in misinformation} is the mean perceived accuracy assigned to false headlines, whereas \textit{news discernment} is the gap between the mean perceived accuracy of true headlines and that of false ones \cite{pennycook2019lazy,calvillo2021personality}. Belief in misinformation is appropriate when a study presents only false content \cite{ahmed2022social,buchanan2021trust}, but news discernment is generally regarded as the more robust indicator because it accounts for judgments across both veridical and deceptive material \cite{guay2023think}. News discernment for user \(k\) is defined as:
\begin{equation}
\textit{ND}_{k} = \frac{1}{n_{\text{true}}} \sum_{i=1}^{n_{\text{true}}} \text{Acc}_{ki}^{\text{true}} - \frac{1}{n_{\text{false}}} \sum_{j=1}^{n_{\text{false}}} \text{Acc}_{kj}^{\text{false}}
\label{eq:discernment}
\end{equation}
where \(\text{Acc}_{ki}^{\text{true}}\) and \(\text{Acc}_{kj}^{\text{false}}\) denote the perceived accuracy ratings assigned by user \(k\) to the \(i\)th true and \(j\)th false headline, respectively. The quantities \(n_{\text{true}}\) and \(n_{\text{false}}\) indicate the total number of true and false headlines rated.

For clarity, we refer to the first term in Equation~\ref{eq:discernment} as the average accuracy rating for real news, denoted \(\text{AR}_{k}\), and to the second term as the average rating for false news, denoted \(\text{AF}_{k}\) (i.e., \manu{the measure of the \textit{belief in misinformation}}). Thus, \(\text{ND}_{k} = \text{AR}_{k} - \text{AF}_{k}\).

Both metrics employ a perceived-accuracy prompt—typically, “To the best of your knowledge, is this headline accurate?”, a question format widely used to probe cognitive and personality correlates of misinformation susceptibility \cite{pennycook2019lazy}. 




\paragraph{Datasets and Materials.}

Our experiments are based on datasets from two previously published studies: \citet{calvillo2021personality} and \citet{huang2024designing}. Calvillo et al.\ \cite{calvillo2021personality} considers a sample of 336 US-based Mechanical Turk workers, including 168 individuals who identified as female, 167 as male, and one who declined to report their gender. Participants ranged in age from 19 to 78 years (\( \textit{Mdn} = 37 \)).
Huang et al.\ \cite{huang2024designing} repurposed data originally collected by \citet{soto2017next}. This sample includes 438 US-based participants: 300 identified as female, 133 as male, and 5 do not disclose their gender. Participant ages range from 16 to 49 years (\( \textit{Mdn} = 21 \)).

Calvillo's dataset \cite{calvillo2021personality} contains the full set of participants' responses to both the personality inventory and the headline evaluation task, as well as the full set of rated headlines. To assess their personality traits according to the Big Five model\footnote{See \url{https://www.colby.edu/wp-content/uploads/2013/08/bfi2s-form.pdf} for the full inventory and scoring procedure.}, participants in \cite{calvillo2021personality} completed the Big Five Inventory–2–Short Form (BFI-2-S; \citealp{soto2017next}): a set of 30 statements rated on a 5-point Likert scale, ranging from 1 (strongly disagree) to 5 (strongly agree). 

For the headline evaluation task, each participant rated 24 news headlines - equally balanced between true and false content, and between pro-liberal and pro-conservative perspectives. True headlines were taken from \url{NPR.org}, a non-profit, independent news agency, while false headlines were taken from \url{Snopes.com}. The rating was based on a 4-point scale (1 = not at all accurate; 4 = very accurate). All headlines originally appeared online between January and April 2020. This dataset is our primary reference point for comparing human and synthetic responses in our experiments.

\paragraph{Procedure.}

Using the OpenAI API \textit{gpt-3.5-turbo} and \textit{gpt-4o}, we instantiate 336 independent agents, one for each participant in the Calvillo et al.~\cite{calvillo2021personality} dataset. Following the prompting protocol of \citet{huang2024designing}, we condition each agent with that participant’s Big-Five profile by supplying 30 item-level answers from the BFI-2-S. 
To assess the influence of prompt format on personality assignment, we implement both 5-point Likert scale and  5-point Expanded scale \citet{huang2024designing}. In the Expanded format, the response options are more descriptive. Typically, there are still five categories, but the labels change from 1 (very inaccurate description for me) to 5 (very accurate). We decide to use the Expanded format also because previous results suggest that, for \textit{gpt-3.5-turbo}, the Expanded format produces synthetic agents whose responses are more similar to those of human participants~\cite{huang2024designing}.

The personality-aligned agents then rate the 24 news headlines from Calvillo et al. on a four-point scale (1 = “not at all accurate”, 4 = “very accurate”), exactly replicating the human task.

Furthermore, we generate 438 additional agents using the personality profiles from \citet{huang2024designing}, this time embedding responses from the full 60-item BFI-2 inventory, again using both Likert and Expanded formats.

\paragraph{Experimental Settings.}

To explore the impact of model stochasticity, we test two temperature settings on both \textit{gpt-3.5-turbo} and \textit{gpt-4o} models: \(0.2\), which encourages deterministic behavior, and \(0.7\), which introduces more creativity and variability. 



To ensure replicability and minimize bias in prompt construction, we adopt the methodology introduced by \citet{huang2024designing}, which embeds personality traits directly into prompts based on participants’ responses to personality inventories. Their released code supports prompt generation from BFI-2 responses in both Likert and Expanded formats. However, their implementation does not support the BFI-2-S, the short-form version of the BFI-2 (30 statements instead of 60).
To enable compatibility, we develop and validate a custom procedure that converts BFI-2-S responses into the Expanded format. This step is essential for incorporating the Calvillo et al.~\cite{calvillo2021personality} dataset, which relies on BFI-2-S for personality assessment, unlike the full BFI-2 used in \citet{huang2024designing}.

\section{Experiments}

\paragraph{Comparing LLM responses with and without personality conditioning.}
To assess whether assigning personality traits to LLMs leads to systematically different judgments compared to standard (unconditioned) LLM behavior, we conduct a comparative analysis across the set of 24 headlines proposed by Calvillo et al. \cite{calvillo2021personality}. 
For each headline, we collect perceived accuracy scores from LLMs prompted with personality-specific instructions, and compare these with responses from the same models prompted without assigning personality traits.

To test whether personality conditioning produces statistically reliable shifts in headline-accuracy ratings, we apply three complementary statistics. First, a Kolmogorov–Smirnov (KS) test compares the overall shapes of the two score distributions (personality-conditioned vs.\ neutral). Second, a Mann–Whitney U (MW) test examines whether the two groups differ in central tendency without assuming normality. Finally, we compute Cohen’s \(d\) to express the magnitude of the mean difference in standard-deviation units, classifying the effect as small, medium, or large.



Because the LLM without personality conditioning produces only one accuracy score, we approximate its sampling distribution by adding zero-mean Gaussian noise to that single value. This synthetic spread yields a neutral response distribution that can be compared directly with the full distributions generated by the personality-conditioned agents.

Table~\ref{tab:llm_response_comparison} summarizes the number of headlines for which statistically significant differences are observed under the KS and MW tests, as well as the number of headlines that fall into different ranges of Cohen’s $d$ effect size. A consistent pattern emerges across configurations: the vast majority of settings yield statistically significant differences for a substantial number of headlines, with many comparisons showing moderate to strong effect sizes. This suggests that conditioning LLMs with personality traits not only shifts their average response levels, but can also alter the distributional properties of their judgments.

\begin{table}[ht]
\centering
\caption{Statistical comparison of LLM responses with and without personality conditioning across models and datasets. ``KS'' and ``MW'' denote the number of headlines with significant differences ($p < .05$) based on Kolmogorov--Smirnov and Mann--Whitney U tests. The others columns show the distribution of effect sizes by Cohen’s $d$ ranges.
}
\label{tab:llm_response_comparison}
\resizebox{\columnwidth}{!}{%
\begin{tabular}{llcccccccc}
\toprule
\textbf{Model} & \textbf{Scale} & \textbf{Temp} & \textbf{KS} & \textbf{MW} & \textbf{d $\leq$ .2} & \textbf{.21–.5} & \textbf{.51–.8} & \textbf{$>$ .8} \\
\midrule

\multicolumn{9}{l}{\small\textit{LLM-based agents using personality profiles from Calvillo et al.\ \cite{calvillo2021personality} (BFI2-S)}} \\
\midrule
GPT-3.5 & Likert   & 0.2 & 24 & 20 & 7 & 2 & 5 & 10 \\
GPT-3.5 & Expanded & 0.2 & 24 & 21 & 2 & 3 & 1 & 15 \\
GPT-3.5 & Likert   & 0.7 & 24 & 20 & 7 & 2 & 5 & 10 \\
GPT-3.5 & Expanded & 0.7 & 24 & 20 & 2 & 3 & 1 & 15 \\
\midrule
GPT-4o  & Likert   & 0.2 & 24 & 22 & 5 & 2 & 6 & 11 \\
GPT-4o  & Expanded & 0.2 & 24 & 19 & 5 & 4 & 1 & 14 \\
GPT-4o  & Likert   & 0.7 & 24 & 21 & 5 & 4 & 5 & 10 \\
GPT-4o  & Expanded & 0.7 & 24 & 18 & 5 & 5 & 0 & 13 \\

\midrule
\multicolumn{9}{l}{\small\textit{LLM-based agents using personality profiles from Huang et al.\ \cite{huang2024designing} (BFI2)}} \\
\midrule
GPT-3.5 & Likert   & 0.2 & 24 & 22 & 3 & 7 & 6 & 8 \\
GPT-3.5 & Expanded & 0.2 & 24 & 20 & 3 & 3 & 3 & 12 \\
GPT-3.5 & Likert   & 0.7 & 24 & 21 & 3 & 7 & 6 & 8 \\
GPT-3.5 & Expanded & 0.7 & 24 & 20 & 3 & 2 & 3 & 12 \\
\midrule
GPT-4o  & Likert   & 0.2 & 24 & 21 & 4 & 5 & 0 & 14 \\
GPT-4o  & Expanded & 0.2 & 24 & 18 & 6 & 1 & 1 & 13 \\
GPT-4o  & Likert   & 0.7 & 24 & 20 & 4 & 4 & 0 & 13 \\
GPT-4o  & Expanded & 0.7 & 24 & 17 & 8 & 1 & 1 & 12 \\

\bottomrule
\end{tabular}
}
\end{table}

\paragraph{News Discernment and Humans.} To ground the study empirically, we first replicate the correlation analysis of \citet{calvillo2021personality}, recomputing Pearson coefficients between each Big-Five trait and news discernment (Eq. \ref{eq:discernment}). This reproducibility check generates the five-element reference vector that serves as our human baseline; only after establishing it do we query the LLMs, so any divergence reflects simulation limits rather than analytic drift.

Table \ref{tab:links_misinfo_and_traits_with_humans_calvillo} reports the two-tailed correlations (\(\alpha = 0.05\)): News discernment is positively related to Agreeableness ($p < .001$), Conscientiousness ($p = .026$), and Open-Mindedness ($p < .001$). No significant correlations are found with Extraversion ($p = .271$) or Negative emotionality ($p = .764$). Inter-trait correlations are included for completeness.

\begin{table}[ht]
\caption{\textbf{Replication of human findings using data from Calvillo et al.~\cite{calvillo2021personality}.} Pearson correlation coefficients between news discernment and personality traits. Significance levels: * \(p < .05\), ** \(p < .01\), *** \(p < .001\).
}
\label{tab:links_misinfo_and_traits_with_humans_calvillo}
\centering
\resizebox{\columnwidth}{!}{%
\begin{tabular}{lccccc}
\toprule
 & E & A & C & N & O \\
\midrule
\textit{News Discernment} & -0.06 & \(0.19^{***}\) & \(0.12^{*}\) & -0.02 & \(0.35^{***}\) \\
Extraversion (E) & 1.0 & \(0.26^{***}\) & \(0.41^{***}\) & \(-0.55^{***}\) & \(0.31^{***}\) \\
Agreeableness (A) &  & 1.0 & \(0.48^{***}\) & \(-0.40^{***}\) & \(0.24^{***}\) \\
Conscientiousness (C) &  &  & 1.0 & \(-0.51^{***}\) & \(0.23^{***}\) \\
Negative Emotionality (N) &  &  &  & 1.0 & \(-0.23^{***}\) \\
Open Mindedness (O) &  &  &  &  & 1.0 \\
\bottomrule
\end{tabular}
}
\end{table}

\paragraph{News Discernment and LLM-based Agents.}
To evaluate the ability of LLM-based agents to simulate human personality-driven news discernment, we follow the psychometric framework proposed by \cite{huang2024designing}. Specifically, we instantiate a group of agents that replicate the personality profiles of participants from Calvillo et al. \cite{calvillo2021personality} (see Section~\ref{sec:methodology} for implementation details). Each agent is then asked to rate the same set of 24 news headlines used in the original study \cite{calvillo2021personality}.
For each synthetic participant, we compute the news discernment score as defined in Equation~\ref{eq:discernment}, mirroring the original procedure in \cite{calvillo2021personality}. We then calculate the bivariate correlations between each personality trait and the discernment scores.
The rationale 
is that similarity in the direction and statistical significance of correlations between synthetic and human data provides evidence that LLM-based agents are capable of simulating human personality traits in news discernment.

\begin{figure}[ht]
\centering
\includegraphics[width=\linewidth]{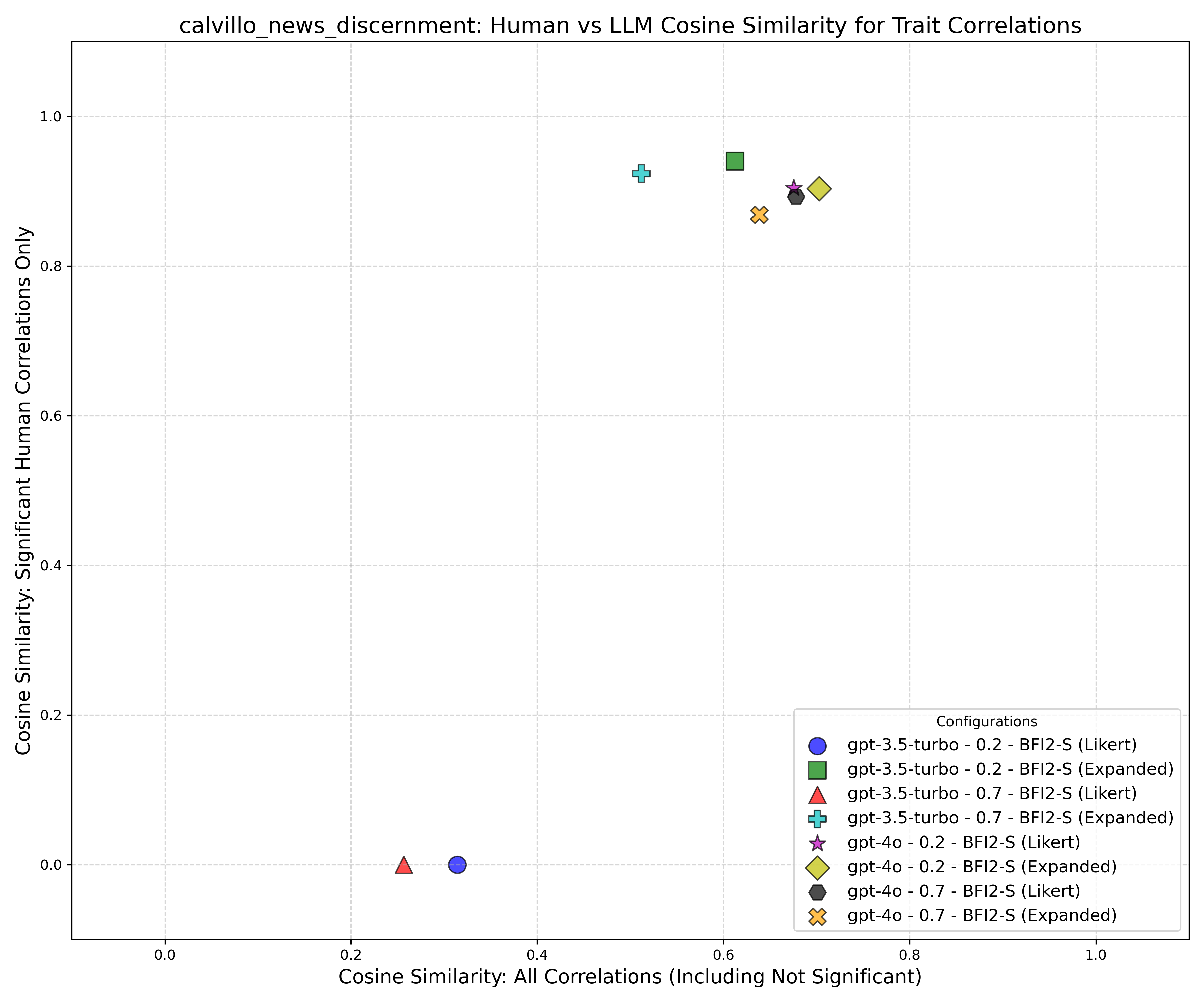}
\caption{\textbf{Similarity between human and LLM-based trait-discernment correlations.} For each model configuration, we report the cosine similarity between the vector of Pearson correlations (personality traits vs. news discernment) and the corresponding human-derived vector from Calvillo et al.~\cite{calvillo2021personality} (see first row of Table~\ref{tab:links_misinfo_and_traits_with_humans_calvillo}). LLM-based agent correlation are produced by simulating profiles derived from the same personality dataset \cite{calvillo2021personality}. The x-axis reports similarity across all traits; the y-axis considers only traits significantly correlated in human data.}
\label{fig:scatter_calvillo_news_disc_similarity}
\end{figure}

As a first exploratory analysis, we compute vectors of bivariate correlations between personality traits and news discernment for each model configuration (i.e., model type, response format (Likert vs. Expanded), and temperature). We then assess the fit of each vector to human observations (Table \ref{tab:links_misinfo_and_simu_traits_with_llm_calvillo}, first row) by calculating the cosine similarity.
Figure~\ref{fig:scatter_calvillo_news_disc_similarity} reports this comparison: the x-axis shows cosine similarity using, for each vector, all correlation coefficients (regardless of significance), while the y-axis considers only those correlations that were statistically significant.

As shown, most configurations, particularly those using GPT-4o, achieve high cosine similarity 
 (between 0.87 and 0.94) when focusing on the significant links between personality traits and news discernment observed in humans (see y-axis values). 
 When non-significant links are considered, similarity scores decrease moderately  (0.51 to 0.70), but still suggest partial alignment. In contrast, the \textit{gpt-3.5-turbo} model with Likert format consistently underperforms, a trend also observed in other evaluation settings \cite{huang2024designing}. Temperature settings appear to have minimal effect on cosine similarity.


Figure \ref{fig:scatter_calvillo_news_disc_similarity} visualizes the alignment between human and synthetic trait effects on news discernment (cosine similarity of their correlation vectors). Table \ref{tab:links_misinfo_and_simu_traits_with_llm_calvillo} reports the corresponding numeric values for every LLM configuration, alongside the human baseline in the row “Calvillo et al.\ \cite{calvillo2021personality} (reference)”. This analysis provides an initial diagnostic for detecting convergence or divergence in bivariate associations between specific personality traits and news discernment.

\begin{table}[ht]


\caption{Pearson correlations between \textit{news discernment} and personality traits across models (using personality profiles from Calvillo et al.~\cite{calvillo2021personality}). Significance: *\(p < .05\), **\(p < .01\), ***\(p < .001\).}
\label{tab:links_misinfo_and_simu_traits_with_llm_calvillo}
\centering
\resizebox{\columnwidth}{!}{%
\begin{tabular}{lccccc}
\toprule
Setting & E & A & C & N & O \\
\midrule
Calvillo et al. \cite{calvillo2021personality} (\textit{reference}) & -0.06 & \(0.19^{***}\) & \(0.12^{*}\) & -0.02 & \(0.35^{***}\) \\
\midrule
GPT-3.5-turbo — 0.2 — BFI2-S (Likert) & \(0.12^{*}\) & 0.02 & 0.07 & 0.05 & 0.05 \\
GPT-3.5-turbo — 0.2 — BFI2-S (Expanded) & \(0.23^{***}\) & \(0.21^{***}\) & 0.11 & \(-0.19^{***}\) & \(0.19^{***}\) \\
GPT-3.5-turbo — 0.7 — BFI2-S (Likert) & \(0.13^{*}\) & 0.04 & 0.01 & 0.03 & 0.04 \\
GPT-3.5-turbo — 0.7 — BFI2-S (Expanded) & \(0.24^{***}\) & \(0.19^{***}\) & 0.09 & \(-0.21^{***}\) & \(0.15^{**}\) \\
GPT-4o — 0.2 — BFI2-S (Likert) & \(0.45^{***}\) & \(0.56^{***}\) & \(0.57^{***}\) & \(-0.59^{***}\) & \(0.53^{***}\) \\
GPT-4o — 0.2 — BFI2-S (Expanded) & \(0.35^{***}\) & \(0.49^{***}\) & \(0.55^{***}\) & \(-0.52^{***}\) & \(0.50^{***}\) \\
GPT-4o — 0.7 — BFI2-S (Likert) & \(0.44^{***}\) & \(0.59^{***}\) & \(0.58^{***}\) & \(-0.56^{***}\) & \(0.51^{***}\) \\
GPT-4o — 0.7 — BFI2-S (Expanded) & \(0.42^{***}\) & \(0.41^{***}\) & \(0.61^{***}\) & \(-0.55^{***}\) & \(0.46^{***}\) \\
\bottomrule
\end{tabular}
}
\end{table}

As also observed in Figure \ref{fig:scatter_calvillo_news_disc_similarity}, GPT-4o agents show a stronger and more consistent alignment with the human personality-news discernment pattern across configurations
 than GPT-3.5 agents, particularly in the Expanded format: Agreeableness, Conscientiousness, and Open-mindedness appear to be aligned in direction and significance with humans. 
 However, there are differences, particularly for traits such as Extraversion and Negative Emotionality, which were not significantly associated with news discernment in humans but show significant effects in synthetic agents. This divergence may indicate that LLMs do capture part of the psychological signal, but still imprint model-specific biases that prevent a full replication of human news-judgment behavior.


\begin{table*}[ht]
\caption{\textbf{Multiple regression}. Ordinary Least Squares (OLS) coefficients from regressions predicting news discernment, real news accuracy, and false news accuracy based on personality traits across models (using personality profiles from Calvillo et al.~\cite{calvillo2021personality}). Significance: *\(p < .05\), **\(p < .01\), ***\(p < .001\).}
\label{tab:multi_reg_news_discernment}
\centering
\resizebox{\textwidth}{!}{%
\begin{tabular}{ll|ccccc|ccccc|ccccc}
\toprule
\multirow{2}{*}{Setting} & \multirow{2}{*}{Scale} & \multicolumn{5}{c|}{\textit{News Discernment (ND)}} & \multicolumn{5}{c|}{\textit{Perceived Accuracy of Real News (AR)}} & \multicolumn{5}{c}{\textit{Perceived Accuracy of False News (AF)}} \\
\cmidrule(lr){3-7} \cmidrule(lr){8-12} \cmidrule(lr){13-17}
& & E & A & C & N & O
 & E & A & C & N & O
 & E & A & C & N & O \\
\midrule
\multicolumn{2}{l|}{Calvillo et al. \cite{calvillo2021personality} (\textit{reference})} & \(-0.13^{***}\) & \(0.09^{*}\) & \(0.05\) & \(0.02\) & \(0.24^{***}\) & \(-0.00\) & \(-0.02\) & \(-0.03\) & \(0.00\) & \(0.08^{**}\) & \(0.13^{***}\) & \(-0.10^{**}\) & \(-0.08^{*}\) & \(-0.02\) & \(-0.16^{***}\) \\
\midrule
GPT-3.5 — 0.2 & Likert & \(0.10^{**}\) & \(0.01\) & \(0.04\) & \(0.09^{**}\) & \(0.01\) & \(0.15^{***}\) & \(-0.32^{***}\) & \(-0.15^{**}\) & \(0.19^{***}\) & \(-0.07\) & \(0.06\) & \(-0.33^{***}\) & \(-0.19^{***}\) & \(0.10^{**}\) & \(-0.08^{*}\) \\
GPT-3.5 — 0.2 & Expanded & \(0.06^{*}\) & \(0.07^{**}\) & \(-0.04\) & \(-0.02\) & \(0.04\) & \(0.15^{***}\) & \(0.08^{*}\) & \(-0.08^{*}\) & \(-0.03\) & \(0.04\) & \(0.09^{***}\) & \(0.01\) & \(-0.04^{***}\) & \(-0.01\) & \(0.00\) \\
GPT-3.5 — 0.7 & Likert & \(0.09^{**}\) & \(0.03\) & \(-0.01\) & \(0.06^{*}\) & \(0.00\) & \(0.13^{**}\) & \(-0.29^{***}\) & \(-0.21^{***}\) & \(0.15^{***}\) & \(-0.08^{*}\) & \(0.04\) & \(-0.32^{***}\) & \(-0.19^{***}\) & \(0.09^{**}\) & \(-0.08^{*}\) \\
GPT-3.5 — 0.7 & Expanded & \(0.06^{**}\) & \(0.06^{*}\) & \(-0.05\) & \(-0.03\) & \(0.03\) & \(0.14^{***}\) & \(0.06^{*}\) & \(-0.08^{**}\) & \(-0.03\) & \(0.02\) & \(0.08^{***}\) & \(0.01\) & \(-0.04^{**}\) & \(-0.00\) & \(-0.00\) \\
\midrule
GPT-4o — 0.2 & Likert & \(0.01\) & \(0.11^{***}\) & \(0.08^{***}\) & \(-0.10^{***}\) & \(0.14^{***}\) & \(0.05^{***}\) & \(-0.00\) & \(-0.01\) & \(0.07^{***}\) & \(0.13^{***}\) & \(0.04\) & \(-0.11^{***}\) & \(-0.09^{***}\) & \(0.17^{***}\) & \(-0.01\) \\
GPT-4o — 0.2 & Expanded & \(-0.02\) & \(0.06^{***}\) & \(0.09^{***}\) & \(-0.07^{***}\) & \(0.11^{***}\) & \(0.03^{*}\) & \(-0.01\) & \(0.09^{***}\) & \(0.05^{***}\) & \(0.10^{***}\) & \(0.05^{**}\) & \(-0.07^{***}\) & \(0.00\) & \(0.11^{***}\) & \(-0.01\) \\
GPT-4o — 0.7 & Likert & \(0.01\) & \(0.12^{***}\) & \(0.09^{***}\) & \(-0.08^{***}\) & \(0.13^{***}\) & \(0.06^{***}\) & \(0.04^{*}\) & \(-0.00\) & \(0.10^{***}\) & \(0.12^{***}\) & \(0.05^{*}\) & \(-0.09^{***}\) & \(-0.09^{***}\) & \(0.18^{***}\) & \(-0.02\) \\
GPT-4o — 0.7 & Expanded & \(0.00\) & \(0.02\) & \(0.12^{***}\) & \(-0.07^{***}\) & \(0.10^{***}\) & \(0.03^{*}\) & \(-0.01\) & \(0.09^{***}\) & \(0.05^{***}\) & \(0.10^{***}\) & \(0.03\) & \(-0.03\) & \(-0.03\) & \(0.12^{***}\) & \(0.00\) \\
\bottomrule
\end{tabular}
}
\end{table*}


Similarly to \cite{calvillo2021personality}, we also conduct a multiple regression analysis to simultaneously examine the relationship between news discernment and the Big Five personality traits in both human participants and LLM-based agents with assigned personalities\footnote{We analyze the data with Pearson correlations and multiple regression. Pearson’s \(r\in[-1,1]\) captures the bivariate link between one trait and news discernment, without adjusting for any other influences. Multiple regression enters all traits simultaneously, so each coefficient reflects a trait’s unique contribution while the others are held constant. In short, correlation is symmetric and uncontrolled, whereas regression is directional and adjusts for covariates.
}. The results are presented in Table \ref{tab:multi_reg_news_discernment} \manu{(see the Supplementary Material for the full set of results)}.
To further investigate whether GPT-based agents replicate human-like psychological patterns, we perform a trait-by-trait comparison of regression results between the language models and the reference human data reported by Calvillo. Since news discernment (ND) is defined as the difference between the perceived accuracy of real news (AR) and the perceived accuracy of false news (AF) (\manu{see equation \ref{eq:discernment}}), we also (i) report the results of separate multiple regression analyses for AR and AF in Table \ref{tab:multi_reg_news_discernment}, and (ii) interpret ND by jointly considering the outcomes of the AR and AF regressions.

\textit{Extraversion.} 
In the human data, Extraversion was negatively related to news discernment ($\beta = -0.13^{**}$). GPT-3.5 and GPT-4o agents, however, showed either positive or null effects. The discrepancy stems from the two components of the ND score—accuracy for real headlines (AR) and for false headlines (AF). Among humans, Extraversion had no effect on AR but a positive effect on AF ($\beta = 0.13^{**}$); more extraverted participants thus are more likely to rate false news as accurate, lowering their overall discernment. GPT agents replicated the positive Extraversion–AF link, albeit more weakly, yet, unlike humans, extraverted LLM agents also rated real news as more accurate, offsetting the discernment loss. (e.g., GPT-4o: $\beta = 0.05^{***}$).
This mismatch in AR predictions explains the overall reversal of the ND relationship in the models.

\textit{Agreeableness.} Human data showed a weak but positive effect of Agreeableness on ND ($\beta = 0.09^{*}$), mainly driven by a negative association with AF ($\beta = -0.10^{**}$), while AR remained negligible. GPT-4o largely captures this dynamic: it also showed a negative link with AF (e.g., $\beta = -0.09^{***}$), and little effect on AR. As a result, GPT-4o reproduces the human-like positive ND effect for Agreeableness, particularly in the Likert setting. 

\textit{Conscientiousness.} In humans, Conscientiousness was only weakly associated with ND ($\beta = 0.05$, n.s.), due to opposite or minimal effects on AR and AF. Interestingly, GPT-based agents show stronger and more consistent patterns: they replicate the negative association between Conscientiousness and AF (e.g., GPT-3.5 Likert: $\beta = -0.19^{***}$) and in some configurations introduce a positive effect on AR. The synthetic Conscientiousness trait results in a more pronounced ND association than that observed in humans, suggesting that GPT models may over-encode an idealized form of conscientious behavior: heightened accuracy and greater resistance to falsehood, compared with real participants.

\textit{Negative Emotionality.} Human data did not show a meaningful relationship between Negative Emotionality and ND or its components. In contrast, GPT models - particularly GPT-4o - show a reduced ability to discriminate between true and false news when simulating individuals high in Negative Emotionality 
(i.e., $-0.10^{***} <= \beta <= -0.07^{***}$), explained by consistent positive links with both AF and AR. 

\textit{Open-Mindedness.} 
In Calvillo et al. \cite{calvillo2021personality}, Open-Mindedness showed a strong positive association with ND ($\beta = 0.24^{***}$), driven by increased AR and reduced AF. GPT-3.5 and GPT-4o successfully replicate this pattern, although with slightly reduced effect sizes (e.g., ND $\beta$ between $0.10^{***}$ and $0.14^{***}$). The models also capture the 
positive association with AR and negative association with AF, although the latter is often smaller or non-significant than for humans.

In summary, consistent with the patterns observed in Table~\ref{tab:links_misinfo_and_simu_traits_with_llm_calvillo}, LLMs appear to be able to simulate human-like news discernment for the traits of Agreeableness (A), Conscientiousness (C), and Open-Mindedness (O), albeit with varying degrees of alignment. In contrast, we observe only limited alignment for Extraversion (E) and Negative Emotionality (N). 

To test robustness, we reran the entire pipeline with the participant profiles from \citet{huang2024designing}, which were collected with the full BFI-2 in both Likert and Expanded formats. Although these profiles differ from the BFI-2-S data used in the Calvillo set, the replicated analyses yielded essentially the same trait–discernment patterns, underscoring the generality of our findings. Full replication results are reported in the Supplementary Materials.

\paragraph{Focusing on Perceived Accuracy of False News.}

As discussed in Section~\ref{sec:methodology}, susceptibility to misinformation is typically assessed through the perceived accuracy of false headlines or the more informative \textit{news discernment} metric, which includes both true and false headlines. However, to allow for a broader comparison with previous findings, we follow~\cite{calvillo2024personality} and report results focusing specifically on the perceived accuracy of false headlines by LLMs.

\begin{table}[ht]
\centering
\caption{\manu{\textbf{Directions of Pearson correlation coefficients between perceived accuracy of false headlines and personality traits}. 'neg', 'pos', 'ns' and '--' denote 'negative', 'positive', 'non-significant correlation' and 'trait was not assessed', respectively; In our LLM-based agents results, a correlation is marked as significant if at least one configuration yields a statistically significant result.
}}

\label{tab:big5_misinfo_summary}
\resizebox{\columnwidth}{!}{%
\begin{tabular}{lllccccc}
\toprule
\textbf{Study} & \textbf{Inventory} & \textbf{Scale} & \textbf{E} & \textbf{A} & \textbf{C} & \textbf{N} & \textbf{O} \\
\midrule
\multicolumn{8}{l}{\small\textit{LLM-based agents \textbf{simulating} personality profiles derived from participants in Calvillo et al.~\cite{calvillo2021personality}
 }} \\
\midrule
GPT-3.5 & BFI2-S & Likert & neg & neg & neg & pos & neg \\
GPT-3.5 & BFI2-S & Expanded & pos & ns & ns & neg & pos \\
GPT-4o & BFI2-S & Likert & neg & neg & neg & pos & neg \\
GPT-4o & BFI2-S & Expanded & neg & neg & neg & pos & neg \\
\midrule
\multicolumn{8}{l}{\small\textit{LLM-based agents \textbf{simulating} personality profiles derived from participants in Huang et al.\ \cite{huang2024designing}}} \\
\midrule
GPT-3.5 & BFI2 & Likert & pos & neg & neg & pos & neg \\
GPT-3.5 & BFI2 & Expanded & pos & pos & pos & neg & pos \\
GPT-4o & BFI2 & Likert & ns & neg & neg & pos & neg \\
GPT-4o & BFI2 & Expanded & pos & neg & ns & pos & ns \\
\midrule
\multicolumn{8}{l}{\small\textit{Studies exploring human belief in misinformation}} \\
\midrule
Calvillo et al. \cite{calvillo2021personality} (\textit{reference}) & BFI2-S & Likert & ns & neg & neg & ns & neg \\
Sindermann et al. \cite{sindermann2021evaluation} & BFI2 & Likert & pos & ns & ns & ns & ns \\
Ahmed \& Rasul \cite{ahmed2022social} & BFI-10 & Likert & pos & ns & neg & pos & pos \\
Ahmed \& Tan \cite{ahmed2022personality} & TIPI & Likert & pos & ns & ns & ns & ns \\
Buchanan \cite{buchanan2021trust} & IPIP & Likert & ns & ns & ns & ns & ns \\
Shephard et al. \cite{shephard2023everyday} & IPIP & Likert & -- & -- & -- & ns & -- \\
\bottomrule
\end{tabular}
}
\end{table}


Table~\ref{tab:big5_misinfo_summary} presents a summary of our findings with GPT (top sections), based on personality data from Calvillo et al.~\cite{calvillo2021personality} and Huang et al.~\cite{huang2024designing}, alongside previously reported results from studies with human participants (bottom section). Full results are provided in the Supplementary Materials. The results in Table~\ref{tab:big5_misinfo_summary} can be seen as an extension of the analysis originally reported in~\cite{calvillo2024personality}.

Also in this case, due to shared inventories and headlines, we use \citet{calvillo2021personality} as our primary reference. Our simulated agents show strong alignment with Calvillo’s findings for Agreeableness (A), Conscientiousness (C) and Open Mindedness (O) which are consistently associated with reduced belief in  misinformation.

For Extraversion (E), Calvillo et al.\ found no significant association, whereas our results are mixed—\manu{depending on personality data we use, scale and model}. For Negative Emotionality (N) while Calvillo et al.\ reported no effect, our simulations consistently indicate a significant \textit{positive} relationship (the more you have this trait, the more likely you are to believe false headlines.) 

Although there are some differences between the correlations observed in the synthetic responses and those reported in the reference study by Calvillo et al.\cite{calvillo2021personality}, we find that the overall patterns are consistent with those identified in other human-centered studies. For example, \citet{ahmed2022social} report positive associations for both Negative Emotionality and Extraversion, and \citet{sindermann2021evaluation} observe similar directional effects for Extraversion.

To explore these patterns further, we also compare multiple regression analysis of our simulation with the Calvillo reference (see \textit{Perceived Accuracy of False News} in Table~\ref{tab:multi_reg_news_discernment} ). Our simulations reproduce several key effects seen in Calvillo et al.'s human data, including negative associations with Conscientiousness and Open Mindedness. The GPT-4o models also replicate the negative association with Agreeableness. Differences emerge for Extraversion and especially for Negative Emotionality, which in our models show inconsistent or even opposite effects depending on the configuration - highlighting the nuanced nature of personality-misinformation dynamics.

\section{Discussion}


This study tests whether personality-conditioned LLMs can reproduce human trait effects on misinformation susceptibility. Using participant data from Calvillo et al. \cite{calvillo2021personality} and the personality-alignment method of Huang et al. \cite{huang2024designing}, we instantiated agents mirroring two human personality datasets. These agents rated the accuracy of true and false headlines with the Calvillo–Pennycook protocol~\cite{calvillo2021personality,pennycook2019lazy}. We then asked whether their news discernment (Eq.~\ref{eq:discernment}) and belief in misinformation match the patterns observed in humans.
Our results provide a nuanced answer. 

First, regarding \textbf{RQ1}, \textit{Does LLM personality conditioning matter?},  the answer is \textbf{yes}. Conditioning LLMs with explicit Big-Five personality profiles yields statistically and behaviorally different headline-accuracy ratings compared with unconditioned models; personality prompts therefore exert a meaningful influence on LLM responses.

Second, regarding \textbf{RQ2}, \textit{Do personality-conditioned LLMs reproduce the trait–news
discernment correlations observed in human samples?}, the answer is \textbf{partly.} Our analyses show that LLMs, especially GPT-4o, reproduce the positive associations that Agreeableness, Conscientiousness, and Open-Mindedness show with news discernment in human data. This effect is visible in both bivariate correlations and multivariate regressions. However, discordant patterns emerged for Extraversion and Negative Emotionality: these traits were weak or non-significant predictors for humans, yet became salient (sometimes in the opposite direction) in several LLM configurations. The result suggests that LLMs internalize some plausible psychological regularities but also introduce model-specific biases.


Third, for \textbf{RQ3}, \textit{Do personality-conditioned LLMs  mirror human trait effects on belief in misinformation?}, the answer is \textbf{partly}.
 Across both datasets, the LLMs mirror human trends: Agreeableness and Conscientiousness were negatively associated with belief in misinformation, while Open Mindedness showed variable but generally protective effects. Again, traits such as Extraversion and Negative Emotionality showed inconsistent or even reversed patterns, depending on the model and prompt configuration.

Discrepancies between simulated agents and real people may arise due to various factors, including limitations of the framework proposed by \citet{huang2024designing}, intrinsic biases of LLMs, and the possible inability of psychological frameworks to fully capture the nuances of human behavior.

Although the framework we use \cite{huang2024designing} has proven capable of generating agents with personality traits that can partially replicate human judgment in news discernment and belief in disinformation, we observe notable variability related to prompt design. In particular, differences in the personality inventories used (e.g., BFI-2 vs. \ BFI-2-S) and in the scale formats (e.g., Likert vs. \ Expanded) appear to influence alignment. For example, configurations such as the GPT-3.5 with a Likert scale showed reduced fidelity to human-like patterns of discernment.
Moreover, the omission of key human factors - such as age, education, political ideology, cognitive ability and cultural background - which are known to shape belief formation in real-world contexts \cite{peter2024role, sultan2024susceptibility}, may further exacerbate the mismatch between LLM agents and human behavior.

In terms of model bias, previous work by Salecha et al.~\cite{salecha2024large} highlights a systematic tendency in LLMs towards socially desirable traits, manifested in particular as increased Extraversion and reduced Negative Emotionality. Consistent with these findings, we found that the correlations between the traits Extraversion and Negative Emotionality and news discernment scores in our study differed from those observed in human samples. Such biases may introduce systematic distortions into personality-based responses, for example, by underestimating the complexity of the role of Extraversion in misinformation susceptibility, or by overestimating the vulnerability of high Negative Emotionality profiles, in contrast to human behavioral data.


\section{Conclusion}

In summary, this study delivers two main messages. First, personality conditioning allows current LLMs to approximate several, but not all, trait-based differences in news discernment and belief in misinformation that have been measured in people, suggesting a viable path toward large-scale, low-risk simulation of human variability. Second, the same experiments expose model-specific biases: some traits are over (or under) represented, and the direction of a few associations even reverses relative to human data. Pinpointing the origins of these distortions (be they rooted in pre-training corpora or the prompt framework itself, and devising systematic debiasing methods) remain open research problems.

Despite these caveats, personality-aligned agents could become a practical asset in the fight against misinformation. They make it possible to stress-test fact checking tools, forecast the impact of deceptive campaigns on diverse sub-populations, and pilot educational interventions before rolling them out to real users. 

\paragraph{Limitations.}
This study has some limitations. We relied on the BFI-2 and its 30-item short form, the BFI-2-S; their reduced length and generic Likert anchors can blunt trait resolution and amplify acquiescence bias. Future work could adopt richer inventories—such as the 100-item HEXACO or item-specific versions of the BFI-2—to capture subtler personality facets. 

The analysis rests on two model families (GPT-3.5 and GPT-4o). Expanding the grid to include open-source models and retrieval-augmented architectures would clarify whether the observed biases are idiosyncratic to a single provider or endemic to current generative NLP systems. Likewise, the human benchmarks come from just two datasets collected in Western contexts; broader, cross-cultural samples are needed to rule out cultural overfitting.

We examined the Big Five. Constructs such as cognitive reflection, political ideology, and media literacy are known to modulate misinformation susceptibility and might interact with personality in ways that synthetic agents could either magnify or suppress. Incorporating these variables would create a more ecologically valid test-bed.

Finally, the headlines differed in number, recency, topic, and linguistic complexity. Such heterogeneity can shift baseline accuracy ratings and thus inflate or dampen trait effects.

\section{Ethics Statement}

\mari{Demonstrating that personality-aligned LLM agents can reproduce human patterns of news discernment offers a clear ethical upside: future studies can probe misinformation mechanisms without repeatedly exposing real participants to deceptive content or collecting additional personal data. At the same time, synthetic agents are approximations, not moral stand-ins for humans. Over-interpreting their outputs could lead to flawed policy or product decisions, while releasing personality-conditioned models at scale might enable new forms of micro-targeted persuasion. We also recognize that linking specific traits to higher misinformation susceptibility can stigmatize individuals or pathologize certain personality profiles.  We therefore (i) in case of acceptance, release all prompts, model versions, and evaluation code to foster transparent replication and auditing; 
and (ii) urge future work to pair synthetic-agent findings with minimal confirmatory human samples before drawing actionable conclusions. Continuous bias audits and strict adherence to OECD/UNESCO AI-ethics guidelines, under institutional oversight, remain essential to ensure that LLM-based simulations enhance, rather than erode, societal trust and individual autonomy.}






\bibliography{main}

\end{document}